УДК 04.272.25, 004.421, 004.032.24

# Метод восстановления многомерных временных рядов на основе выявления поведенческих шаблонов и применения автокодировщиков

© 2023 А.А. Юртин

*Южно-Уральский государственный университет*
*(454080 Челябинск, пр. им. В.И. Ленина, д. 76)*
*E-mail: iurtinaa@susu.ru*

**Аннотация**: В данной статье представлен метод для восстановления пропущенных значений в многомерных временных рядах. Метод объединяет технологии нейронных сетей и алгоритм поиска сниппетов (поведенческих шаблонов временного ряда). Он включает этапы предварительной обработки данных, распознавания и реконструкции, применяя сверточные и рекуррентные нейронные сети. Эксперименты показали высокую точность восстановления и преимущество метода перед аналогами.

**Ключевые слова:** временной ряд, восстановление пропущенных значений, автоэнкодеры, нейронная сеть.

## 1 Введение

В современных исследованиях и практических приложениях, таких как мониторинг технического состояния оборудования, управление системами жизнеобеспечения, анализ климатических данных и финансовое прогнозирование, часто возникает задача обработки временных рядов, содержащих пропущенные значения. Эти пропуски могут быть вызваны различными факторами, включая аппаратные сбои и ошибки при сборе данных.

В данной статье представлен метод SAETI (Snippet-based Autoencoder for Time Series Imputation), который является улучшением предыдущей работы авторов [1] и объединяет использование нейронных сетей с концепцией сниппетов для восстановления пропущенных значений временных рядов. В отличие от предыдущей работы, новый метод иллюстрирует альтернативный способ обработки данных, сочетая в себе глубокие нейросети и инновационные техники анализа временных рядов. Метод SAETI предполагает две последовательно применяемые нейросетевые модели: Распознаватель и Реконструктор. Распознаватель получает на входе подпоследовательность координаты ряда, в которой имеются пропущенные точки, и определяет сниппет (поведенческий шаблон) ряда, на который наиболее похожа данная подпоследовательность. Реконструктор, используя входную подпоследовательность и наиболее похожий на нее сниппет, найденный Распознавателем, восстанавливает пропуски. Нахождение сниппетов ряда, подготовка обучающих выборок и обучение указанных выше нейросетевых моделей выполняются в рамках предварительной обработки данных.

Для реализации Реконструктора в новом методе использует модель типа автоэнкодер. Данное изменение делает метод более адаптивным к различным условиям и позволяет восстанавливать не только последние точки, но и пропусками, расположенными случайным образом. Для оценки и сравнения предложенного метода были проведены вычислительные эксперименты, охватывающие наборы данных из различных областей и различные сценарии, приближенные к реальным сценариям формирования пропусков временного ряда.

Данная статья организована следующим образом. Раздел 2 содержит краткий обзор похожих по тематике работ. В разделе 3 приводится список, используемых в статье, определений и нотаций. Раздел 4 содержит подробное описание каждого этапа обработки данных предложенного метода. Результаты вычислительных экспериментов, демонстрирующие качественные характеристики предложенного метода содержатся в Разделе 5. Заключение содержит сводку полученных результатов.

## 2 Обзор связанных работ

В последние годы нейросетевые методы стали занимать ведущие позиции в области восстановления данных временных рядов. Одним из ключевых направлений здесь является использование рекуррентных нейронных сетей (RNN), которые, благодаря своей способности улавливать временные зависимости в данных, находят применение в самых различных задачах. Примером таких моделей является BRITS (Bidirectional Recurrent Imputation for Time Series) [19]. Данная модель применяет двунаправленную рекуррентную сеть для более глубокого анализа последовательностей данных, что позволяет эффективно работать с пропущенными значениями. Особенностью BRITS является её способность одновременно учитывать информацию, как из прошлых, так и из будущих точек временного ряда, что обеспечивает более полное и точное восстановление данных.

Другой группой архитектур, успешно применяемых для восстановления данных временных рядов являются генеративно-состязательные сети (GAN) и автоэнкодеры. Данные модель после обучения способны генерировать высококачественные данные, в следствие обнаружения главных скрытые зависимости в сложных наборах данных.

Примером модели, реализующей архитектуру автоэнкодер является NAOMI (Non-Autoregressive Multiresolution Imputation) [2]. Эта архитектура включает в себя две основные компоненты: Энкодер и Декодер. Энкодер занимается преобразованием входных данных в скрытое состояние с существенно меньшей размерностью по сравнению с исходными данными. Затем это скрытое состояние передается Декодеру, который формирует итоговый ответ.

Примером, реализующим генеративно-состязательные сети для востановления данных временных рядов, является E2GAN (End-to-End Generative Adversarial Network) [3]. E2GAN включает генератор и дискриминатор: генератор использует автоэнкодер и рекуррентные ячейки для сжатия и последующего восстановления временных рядов, а дискриминатор оценивает степень реалистичности восстановленных данных. Части нейронной сети участвуют в состязании, где генератор стремится сформировать убедительные данные, а дискриминатор классифицирует данные на подлинные и сгенерированные.

Еще одной архитектурой, которая может использоваться для анализа последовательных данных является Трансформер [4]. Применение Трансформеров в восстановлении данных временных рядов представляет собой новую и перспективную область исследований. Трансформеры, изначально разработанные для задач обработки естественного языка, показывают обнадеживающие результаты в области анализа временных рядов благодаря своей способности эффективно обрабатывать последовательные данные. Модели, реализующие данную архитектуру, отличаются от традиционных подходов, таких как рекуррентные нейронные сети (RNN), тем, что могут обрабатывать все данные одновременно, а не последовательно. Одним из ключевых компонентов данной системы является механизм Самовнимания. Данный механизм посредством матричных операций над входными данными позволяет модели фокусироваться на разных частях данных, чтобы лучше понять их структуру и взаимосвязи.

Примером успешного использования данной архитектуры к задаче восстановления временных рядов является Модель SAITS (Self-Attention-based Imputation for Time Series) [22]. Эта модель использует механизмы самовнимания для восстановления пропущенных данных во временных рядах. Самовнимание позволяет модели акцентировать внимание на различных частях входной последовательности, что обеспечивает более точное восстановление данных, основываясь на контекстной информации из всего временного ряда

## 3 Основные определения и нотации

Ниже приводятся обозначения и определения, используемые в данной статье терминов в соответствии с работами [5, 6, 7].

## 3.1 Временной ряд и подпоследовательность

*Временной ряд (time series)* $T$ представляет собой набор координат – векторов одинаковой длины. Каждая координата многомерного временного ряда представляет собой последовательность хронологически упорядоченных вещественных значений:

$$T^{(j)} = \left(t_1^{(j)}, \ldots, t_n^{(j)}\right),\ t_i^j \in \mathbb{R}, 1 \leq j \leq d. \qquad (1)$$

Число $n$ обозначается $|T^j|$ и называется длиной ряда. Количество координат временного ряда обозначается как $d$.

*Элемент* $T_i$ представляет вектор, состоящий и $i$ элементов каждой координаты $T$:

$$T_i = \left(t_i^{(1)}, \ldots, t_i^{(d)}\right),\ t_i^j \in \mathbb{R}, 1 \leq i \leq n, 1 \leq j \leq d. \qquad (2)$$

*Подпоследовательность (subsequence)* $T_{i,m}$ временного ряда $T$ представляет собой непрерывный промежуток из $m$ элементов, начиная с позиции $i$:

$$T_{i,m} = (t_i, \ldots, t_{i+m-1}), 1 \leq m \ll n,\ 1 \leq i \leq n - m + 1. \qquad (3)$$

Множество всех подпоследовательностей ряда $T$, имеющих длину $m$, обозначим как $S_T^m$, а мощность такого множества за $N$, $N = |S_T^m| = n - m + 1$.

## 3.2 Сниппеты (типичные подпоследовательности) временного ряда

Для упрощения изложения, в данном разделе под временным рядом $T$ будет пониматься одна из координат временного ряда $T^j$. В дальнейшей работе описанный ниже алгоритм применяется к каждой координате временного ряда независимо.

*Задача поиска типичных подпоследовательностей* временного ряда основана на концепции *сниппетов (snippet)*, которая предложена Кеогом и др. в работе [5] и уточняет понятие типичных подпоследовательностей временного ряда следующим образом. Каждый сниппет представляет собой один из сегментов временного ряда. Со сниппетом ассоциируются его ближайшие соседи – подпоследовательности ряда, имеющие ту же длину, что и сниппет, которые более похожи на данный сниппет, чем на другие сегменты. Для вычисления схожести подпоследовательностей используется специализированная мера схожести, основанная на евклидовом расстоянии. Сниппеты упорядочиваются по убыванию мощности множества своих ближайших соседей. Формальное определение сниппетов выглядит следующим образом.

Временной ряд $T$ может быть логически разбит на *сегменты* – непересекающиеся подпоследовательности заданной длины $m$. Здесь и далее без существенного ограничения общности мы можем считать, что $n$ кратно $m$, поскольку $m \ll$ $n$. Множество сегментов ряда, имеющих длину $m \ll n$, обозначим как $S_T^m$, элементы этого множества как $S_1, \ldots, S_{n/m}$:

$$S_T^m = (S_1, \ldots, S_{n/m}), \text{где } S_i = T_{m \cdot (i-1)+1, m}. \quad (3.2.1)$$

Обозначим множество сниппетов ряда $T$, имеющих длину $m \ll n$, как $C_T^m$, а элементы этого множества – как $C_1, \ldots, C_K$:

$$C_T^m = (C_1, \ldots, C_K), \text{где } C_i \in S_T^m. \quad (3.2.2)$$

Число $K$ ($1 \leq K \leq n/m$) представляет собой параметр, задаваемый прикладным программистом, и отражает соответствующее количество наиболее типичных сниппетов. С каждым сниппетом ассоциированы следующие атрибуты: индекс сниппета, ближайшие соседи и значимость данного сниппета.

*Индекс сниппета* $C_i \in C_T^m$ обозначается как $C_i.index$ и представляет собой номер $j$ сегмента $S_j$, которому соответствует подпоследовательность ряда $T_{m \cdot (j-1)+1, m}$.

*Множество ближайших соседей сниппета* $C_i \in C_T^m$ обозначается как $C_i.Neighbors$ и содержит подпоследовательности ряда, которые более похожи на данный сниппет, чем на другие сегменты ряда, в смысле меры схожести MPdist [23]:

$$C_i.Neighbors = \{T_{j,m} \mid S_{C_i.index} = \arg\min_{1 \leq r \leq n/m} \text{MPdist}(T_{j,m}, S_r), 1 \leq j \leq n-m+1\}. \quad (3.2.3)$$

*Значимость сниппета* $C_i \in C_T^m$ обозначается как $C_i.frac$ представляет собой долю множества ближайших соседей сниппета в общем количестве подпоследовательностей ряда, имеющих длину $m$:

$$C_i.frac = \frac{|C_i.Neighbors|}{n-m+1}. \quad (3.2.4)$$

Сниппеты упорядочиваются по убыванию их значимости:

$$\forall C_i, C_j \in C_T^m: i < j \Leftrightarrow C_i.frac \geq C_j.frac. \quad (3.2.5)$$

Таким образом, рассматриваемая задача состоит в том, чтобы для заданных временного ряда $T$, длины сегмента $m$ и количества значимых сниппетов $K$ найти наиболее значимые сниппеты $C_1, \ldots, C_K \in C_T^m$, в т.ч. индекс, множество ближайших соседей и значимость каждого из упомянутых сниппетов.

# 4 Нейросетевой метод восстановления пропущенных значений на основе автоэнкодера и сниппетов

Метод восстановления пропущенных значений многомерного временного ряда SAETI (Snippet-based Autoencoder for Time Series Imputation), сочетает в себе технологии нейронных сетей и параллельный алгоритм поиска сниппетов. Данный метод предполагает наличие двух нейросетевых моделей: Распознаватель и Реконструктор. Распознаватель получает на входе подпоследовательность ряда, случайные точки, которой пропущены, и определяет сниппеты (типичные подпоследовательности) координат ряда, на которые наиболее похожи координаты данной подпоследовательность. Реконструктор, используя подпоследовательность исходного ряда с пропущенными точками и наиболее похожие на нее сниппеты, найденный Распознавателем, восстанавливает пропущенную точку. Нахождение сниппетов ряда, подготовка обучающих выборок и обучение указанных выше нейросетевых моделей выполняются в рамках предварительной обработки данных.

## 4.1 Предварительная обработка

Во время предварительной обработки данных формируются обучающие выборки для Распознавателя и Реконструктора. Предварительная обработка состоит из следующих шагов: нормализации данных и поиска сниппетов.

Для предварительной обработки выбирается репрезентативный временной ряд $T$ ($|T| = n$), включающий все активности субъекта, описываемого данным рядом. Пусть длина сниппета равна $m$ ($m \ll n$) и количество активностей ($1 < K \ll n/m$).

Во время нормализации ряд $T$ приводится к диапазону $[0, 1]$ по средствам минимаксной нормализации. В нормализованном временном ряде происходит поиск сниппетов. В качестве реализации поиска сниппетов используется алгоритм PSF (Parallel Snippet-Finder) [8]. Перед подачей на вход алгоритма из временного ряда исключаются подпоследовательности, содержащие пропущенные значения. В результате поиска подпоследовательности исходного временного ряда помечаются в соответствии с их принадлежностью активностям временного ряда. Полученная разметка в дальнейшем используется для формирования обучающих выборок Распознавателя. Обучающая выборка Реконструктора формируется из всех подпоследовательностей временного ряда, включая подпоследовательности содержащие пропуски.

## 4.2 Распознаватель

На рис. представлена структура нейронной сети, реализующей Распознаватель. На вход нейронной сети Распознавателя поступает подпоследовательность временного ряда, содержащая пропущенные значения. Пропущенные значения как для Распознавателя, так и для Реконструктора перед подачей на вход моделей заменяются на -1. На выходе нейронной сети для каждой координаты формируется вектор вероятности принадлежности подпоследовательности к множеству ближайших соседей соответствующего сниппета. В дальнейшем Распознаватель выбирает сниппеты для каждой координаты, вероятность принадлежности к которым максимальна.

Данная нейронная сеть состоит из следующих последовательно применяемых трех групп слоев: сверточных, рекуррентного и полносвязного. Сверточные слои отвечают за выделение основных признаков из входной подпоследовательности. Данные слои включают три сверточных слоя с количеством фильтров (ядер) 256, 128 и 64 соответственно. Размер ядра (фильтра) для каждого слоя составляет 5 значений. После каждого сверточного слоя происходит процесс сжатия данных с помощью операции подвыборки по максимальному значению (max-pooling). Сжатие данных происходит путем уменьшения размерности каждой карты признаков предыдущего сверточного слоя. Размер подвыборки составляет 2. В качестве функции активации был выбран Линейный выпрямитель (ReLU, Rectified linear unit) [9]. Рекуррентный слой, состоящий из 128 управляемых рекуррентных блоков (Gated Recurrent Units, GRU), анализирует выделенные предыдущими слоями признаки с учетом временного контекста.

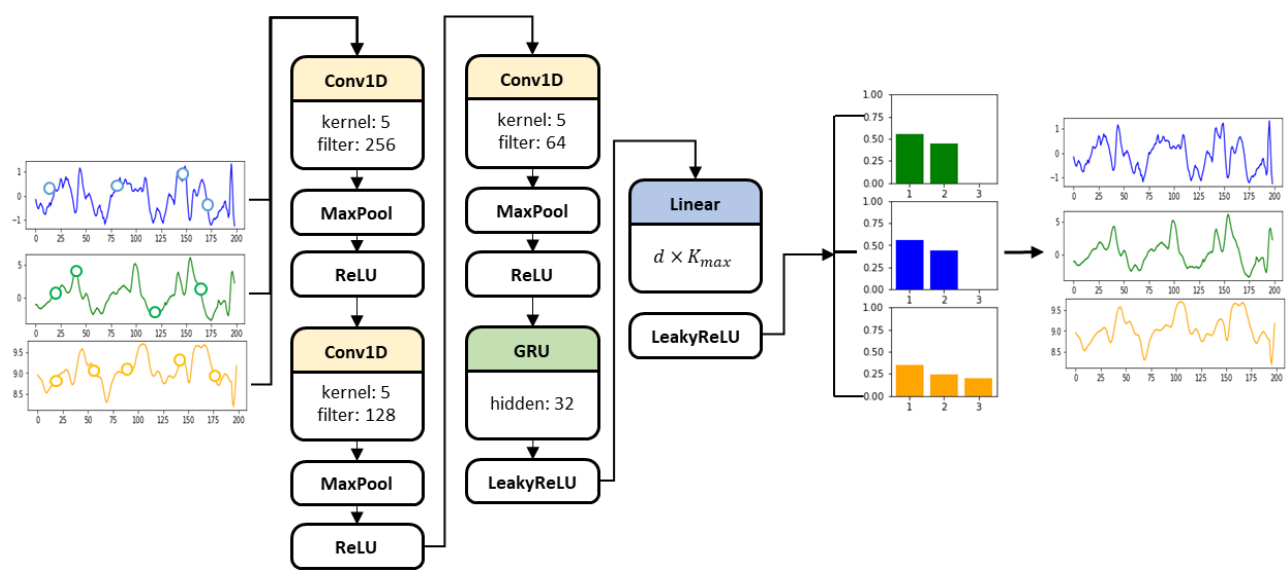

Рис. 1. Структура нейронной сети Распознавателя

В качестве функции активации этого и всех последующих нейронов применяется Линейный выпрямитель с "утечкой" (Leaky ReLU), позволяющего избежать проблемы «умирающего» ReLU для глубоких нейронных сетей [10]. Полносвязный слой, включающий $d \times K$, нейронов вычисляет итоговый вектор вероятностей для каждой координаты входной подпоследовательности. Для формирования итого набора сниппетов, к каждому итоговому вектору применяется операция argmax.

### 4.3 Реконструктор

На рис. 2 представлена структура нейронной сети, реализующей Реконструктор.

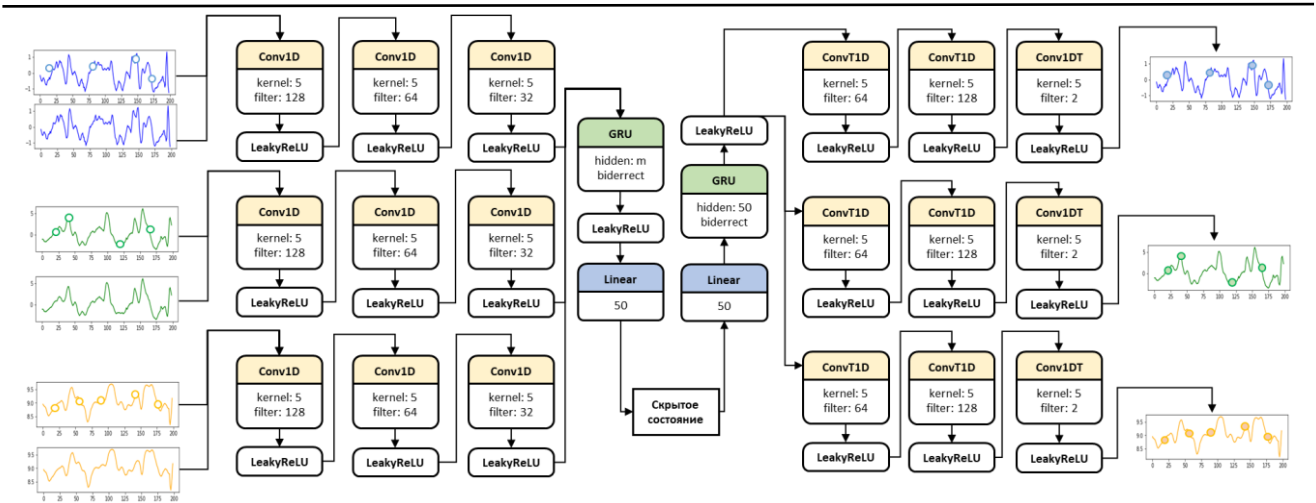

Рис. 2. Структура нейронной сети Реконструктора

На вход данной нейронной сети подается набор матриц, состоящих из двух строк: подпоследовательности временного ряда, в которой пропуски были заменены на -1 и сниппета, предсказанного Распознавателем. На выходе нейронной сети формируется близкая к входной подпоследовательностей, в которой пропущенные точки заменены синтетическими значениями.

Нейронная сеть Реконструктор представляет собой нейронную сеть класса Автоэнкодер [11], включающую две компоненты: Кодировщик (Encoder) и Декодировщик (Decoder). Во время обработки входной подпоследовательности автоэнкодером, кодировщик сжимает данные в скрытое представление, которое является сжатой формой исходных данных. Скрытое представляет собой вектор в пространстве меньшей размерности по сравнению с исходными данными. Данный вектор включает ключевые характеристики и паттерны входных данных, удаляя при этом избыточную или нерелевантную для восстановления информацию. Скрытое представление подается на вход Декодировщика, производящему процесс декодирования данных. В результате формируется подпоследовательность приближенная к входным данным. Кодировщик включает 3 группы слоев: сверточную, рекуррентный и полносвязный. Сверточные слои отвечают за извлечение главных признаков из объединения входной подпоследовательности и сниппета. Слои, представляют набор сверточных слоев для каждой координаты входной подпоследовательность. Каждая последовательность состоит из 3 слоев с 128, 64 и 32 фильтрами (ядрами) соответственно. Размер каждого фильтра (ядра) равен 5. Выходы каждой последовательности слоев сцепляются и направляются на вход общего рекуррентного слоя, отвечающего за анализ данных с учетом временного контекста. Рекуррентный слой состоит из $m$ управляемых рекуррентных блоков (Gated Recurrent Units, GRU). Полносвязный слой отвечает за формирование скрытого представления и состоит из такого числа нейронов, равного размерности скрытого представления. Структура Декодировщика зеркально отражает структуру Кодировщика – скрытое представление поступает на вход полносвязного слоя и рекуррентного слоя после чего разделяются на координаты и обкатываются независимо наборами сверточных слоев.

## 4.4 Обучение модели

Обозначим обучающую выборку нейронной сети как множество пар $D = \{<X;Y>\}$, где $X$ представляет собой входные данные, а $Y$ – соответствующие им выходные данные. Кортеж обучающей выборки для нейронной сети Распознавателя формируется следующим образом. Входным данным полагаются подпоследовательности репрезентативного фрагмента $T$, выходным данным – вектора длины $d$. Каждому значению вектора соответствует номер сниппета, к которому принадлежит координата подпоследовательности. Фрагмент $T$ для обучения Распознавателя разделяется следующим образом. Для обучения распознавателя подпоследовательности не содержащие пропусков фрагмента $T$ случайным образом распределяется на обучающую, валидационную и тестовую выборку. Процентное соотношение получившихся выборок следующие: 75% для обучения модели и 25% для валидации.

Кортеж обучающей выборки Реконструктора формируется следующим образом. Атрибут $X$ представляет собой матрицу из $d \times 2$ строк, где в качестве каждой второй строки подставляются сниппеты, а первой поочередно подпоследовательности временного ряда, включая подпоследовательности содержащие пропуски. Атрибутом $Y$ являются копия атрибута $X$. Здесь важно отметить, что формирование обучающей выборки Реконструктора осуществляется после обучения Распознавателя на подготовленной для него выборке: сниппеты, являющиеся первыми строками матрицы входных данных Реконструктора, представляют собой выход Распознавателя. Во время обучения, 25% точек, подаваемых на вход нейросетевых моделей, случайных образом заменяются пропусками.

### 4.4.1 Восстановление ряда

Работа метода SAETI может быть описана следующим образом. Пусть обучающие выборки Распознавателя и Реконструктора сформированы на основе временного ряда $T$ и указанные нейросетевые модели обучены на этих данных, как описано выше. Пусть задан временной ряд $U$, который отражает активность того же субъекта, что и ряд $T$, однако содержит пропущенные значения. Ряд $U$ нормализуется и разделяется на набор непересекающихся подпоследовательностей. Из полученного набора выделяются все подпоследовательности, содержащие пропуски, и последовательно поступают на вход нейронным сетям метода. Из подпоследовательностей, сформированных Реконструктором, выбираются только те точки, которые были пропущенных и после нормализации подставляются в исходный временной ряд.

# 5 Вычислительные эксперименты

Для исследования эффективности предложенного метода были проведены вычислительные эксперименты на оборудовании Лаборатории суперкомпьютерного моделирования ЮУрГУ [12]. В экспериментах исследовалась точность восстановления предложенного метода в различных предметных областях и сравнивалась с различными передовыми методами восстановления.

## 5.1 Описание экспериментов

Для оценки точности восстановления используется корень среднеквадратичной ошибки (Root Mean Square Error, RMSE):

$$RMSE = \sqrt{\frac{1}{n}\sum_{i=0}^{n}(y_i - \widehat{y_i})^2}, \qquad (4)$$

где $y_i$ – реальные данные, $\widehat{y_i}$ – прогноз модели, $n$ – количество точек.

### 5.1.1 Наборы данных

Для оценки качества предложенного метода и сравнения его с аналогами использовались наборы данных указанные в табл. . Наборы Madrid и Walk Run можно отнести к наборам, содержащим активности. В первом случае активностями могут выступать дни или часть недели с разной активностью проезжающих машин. Для Walk Run такими активностями могут быть различные физические упражнения.

Табл. 1 Наборы данных для восстановления

| № п/п | Временной ряд | Предметная область | Длина ряда, тыс. $n$ | Количество координат, $d$ |
|---|---|---|---|---|
| 1 | BAFU [13] | Сброс воды в реках Швейцарии | 50 | 10 |
| 2 | Climate [13] | Погода в различных местах Сев. Америки | 5 | 10 |
| 3 | Madrid [14] | Дорожный трафик Мадрида | 25 | 10 |
| 4 | Marel [13] | Характеристики морской воды в Ла-Манше | 50 | 10 |
| 5 | Walk Run | Показания нательных датчиков во время физ. активности человека | 220 | 11 |

### 5.1.2 Аналоги

В экспериментах предложенный метод сравнивался с передовыми аналитическими и нейросетевыми методами восстановления. К аналитическим методам относятся следующие: CDRec [15], DynaMMo [16], ROSL [17], SVT [18]. К нейросетевым: BRITS [19], GP-VAE [20], MRNN [21], SAITS и TRANSFORM [22]. В качестве готовой реализации аналитических методов восстановления использовался фреймворк ImputeBench [13].

### 5.1.3 Сценарии

Для формирования тестовых пропусков в исходных рядах использовались сценарии, описанные в ImputeBench [13].

Сценарий Blackout симулирует поведение, когда одновременно все источники данных отключаются, формируют не продолжительные синхронные пропуски во всех координатах временного ряда. Для вычислительных экспериментов длина пропуска была выбрана равной 10 и 100 точек.

Сценарий MCAR симулирует поведение, когда на непродолжительное время отключается один из источников данных, формируя пропуски в одной из координат временного ряда. Во время сценария координата циклически выбирается случайным образом, до тех пор, пока объем пропусков не достигнет 25%.

Сценарий TS NBR симулирует продолжительное отключение одного из источников данных.

## 5.2 Результаты

Табл. 1 Результаты восстановления для сценария Blackout

| № п/п | Метод | RMSE | | | | |
|---|---|---|---|---|---|---|
| | | BAFU | Climate | Madrid | MAREL | Walk Run |
| *Аналитические методы* | | | | | | |
| 1 | CDRec | 0.0054 | 0.17 | 0.1276 | 0.122 | 0.135 |
| 2 | DynaMMo | 0.0389 | 0.2203 | 0.1377 | 0.162 | 0.2329 |
| 3 | ROSL | 0.0453 | 0.1646 | 0.1949 | 0.119 | 0.2066 |
| 4 | SVT | 0.1612 | 0.4272 | 0.2139 | 0.516 | 0.6053 |
| *Нейросетевые методы* | | | | | | |
| 5 | BRITS | 0.07 | 0.17 | 0.13 | 0.32 | 0.39 |
| 6 | GP-VAE | 0.008 | 0.216 | 0.1221 | 0.29 | 0.38 |
| 7 | MRNN | 0.298 | 0.36 | 0.19 | 0.38 | 0.365 |
| 8 | SAITS | 0.03 | 0.09 | 0.117 | 0.114 | 0.16 |
| 9 | TRANSFORM | 0.03 | 0.07 | 0.1 | 0.14 | 0.17 |
| 10 | **SAETI** | 0.005 | 0.16 | 0.0304 | 0.08 | 0.06 |

В табл. *1*2, табл. 2 и табл. *3*4 **Ошибка! Источник ссылки не найден.**представлены вычислительные эксперименты для всех наборов и на всех методах для сценариев blackout, MCAR и TS NBR соответственно. Эксперименты демонстрируют, что предложенный метод превосходит точность конкурентов на наборах данных, содержавшие активности временного ряда. В случае если в наборе отсутствуют активности предложенный метод в основном входит в топ 5 лучших методов восстановления. На рис. 2представлен пример восстановления данных на примере набора данных Madrid для сценария blackout. Как видно из рисунка, предложенный метод распознает не только общие паттерны поведения трафика

Мадрида, но и использует сниппеты для уточнения деталей для конкретных дней-активностей.

Табл. 2 Результаты восстановления для сценария MCAR

| № п/п | Метод | RMSE | | | | |
|---|---|---|---|---|---|---|
| | | BAFU | Climate | Madrid | MAREL | Walk Run |
| | *Аналитические методы* | | | | | |
| 1 | CDRec | 0.0617 | 0.2061 | 0.1366 | 0.2995 | 0.144 |
| 2 | DynaMMo | 0.044 | 0.1726 | 0.09 | 0.128 | 0.157 |
| 3 | ROSL | 0.0508 | 0.1903 | 0.1113 | 0.1895 | 0.115 |
| 4 | SVT | 0.0565 | 0.1847 | 0.1006 | 0.2011 | 0.1307 |
| | *Нейросетевые методы* | | | | | |
| 5 | BRITS | 0.1 | 0.07 | 0.134 | 0.38 | 0.186 |
| 6 | GP-VAE | 0.04 | 0.2628 | 0.07 | 0.092 | 0.081 |
| 7 | MRNN | 0.153 | 0.24 | 0.18 | 0.27 | 0.42 |
| 8 | SAITS | 0.02 | 0.06 | 0.05 | 0.06 | 0.18 |
| 9 | TRANSFORM | 0.04 | 0.04 | 0.06 | 0.1 | 0.083 |
| 10 | **SAETI** | 0.03 | 0.21 | 0.032 | 0.12 | 0.0763 |

Табл. 3 Результаты восстановления для сценария TS NBR

| № п/п | Метод | RMSE | | | | |
|---|---|---|---|---|---|---|
| | | BAFU | Climate | Madrid | MAREL | Walk Run |
| | *Аналитические методы* | | | | | |
| 1 | CDRec | 0.043 | 0.2038 | 0.2226 | 0.3245 | 0.125 |
| 2 | DynaMMo | 0.091 | 0.1678 | 0.168 | 0.264 | 0.32 |
| 3 | ROSL | 0.0292 | 0.233 | 0.1553 | 0.3317 | 0.0908 |
| 4 | SVT | 0.0299 | 0.2288 | 0.1277 | 0.2146 | 0.0869 |
| | *Нейросетевые методы* | | | | | |
| 5 | BRITS | 0.059 | 0.22 | 0.2 | 0.44 | 0.147 |
| 6 | GP-VAE | 0.283 | 0.356 | 0.183 | 0.293 | 0.233 |
| 7 | MRNN | 0.29 | 0.32 | 0.2 | 0.48 | 0.432 |
| 8 | SAITS | 0.7 | 0.04 | 0.17 | 1.7 | 0.147 |
| 9 | TRANSFORM | 0.3 | 0.03 | 0.21 | 1.9 | 0.218 |
| 10 | **SAETI** | 0.112 | 0.14 | 0.138 | 0.3 | 0.0445 |

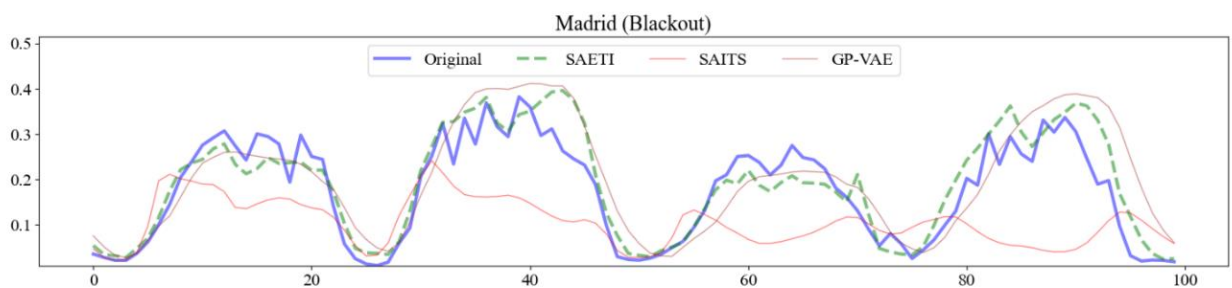

Рис. 2. Пример восстановления данных

# 6 Заключение

В данной статье исследуется важная проблема восстановления пропущенных значений во временных рядах, актуальная для многих прикладных задач. Представлен метод SAETI, основанный на сочетании нейросетевых подходов с алгоритмами поиска сниппетов. Он включает этапы предварительной обработки данных, распознавания и реконструкции с использованием сверточных и рекуррентных нейросетевых слоев, а также автоэнкодера, обеспечивающего эффективное восстановление пропущенных значений. Предварительная обработка данных направлена на построение обучающих выборок для моделей Распознавателя и Реконструктора, реализующие этап распознавания и реконструкции соответственно.

Нейросетевая модель Распознавателя представляет собой многоклассовый классификатор подпоследовательностей, содержащих пропуски, где классами полагаются активности, соответствующие найденным сниппетам (поведенческим шаблонам).

Реконструктор представляет собой нейронную сеть класса автоэнкодер, которая состоит из двух компонентов: Кодировщик (encoder) и Декодировщик (decoder). Кодировщик получает на вход конкатенацию сниппетов и восстанавливаемой подпоследовательности (по каждой координате временного ряда) и формирует скрытое представление данных, которое сохраняет в себе ключевые характеристики входных данных, достаточные для их корректного декодирования. Декодировщик, используя скрытое представление, формирует подпоследовательность, близкую к входным данным, в которой пропуски заменяются синтетическими значениями.

Результаты экспериментов подтверждают высокую эффективность метода SAETI и демонстрируют, что SAETI входи в топ 3 лучших метода восстановления среди лучших аналогичных решений в области анализа временных рядов, содержащих различные активности.

# Список литературы

## Информация об авторах

*Алексей Артемьевич Юртин* — программист лаборатории больших данных и машинного обучения; Южно-Уральский государственный университет (национальный исследовательский университет), пр-т им. В. И. Ленина, д. 76, 454080, Челябинск, Российская Федерация